\documentclass[conference]{IEEEtran}
\IEEEoverridecommandlockouts
\usepackage{amsmath,amssymb}
\usepackage{graphicx}
\usepackage{booktabs}
\usepackage{algorithm}
\usepackage{algpseudocode}
\usepackage{url}
\usepackage{cite}
\usepackage{xcolor}

\begin{document}
\title{Differentially Private Federated Class-Incremental Learning for EEG Mental State Classification\thanks{Accepted at the TrustFL-SIS: Trustworthy Federated Learning for Smart Industrial Systems Workshop, IEEE International Conference on Data Mining (ICDM 2026).}}

\author{
\IEEEauthorblockN{Baran Can G\"ul, Ali G\"ul\"umser, Nasser Jazdi, Michael Weyrich}
\IEEEauthorblockA{Institute of Industrial Automation and Software Engineering\\
University of Stuttgart, Pfaffenwaldring 47, 70550 Stuttgart, Germany\\
\{baran-can.guel, nasser.jazdi, michael.weyrich\}@ias.uni-stuttgart.de, st186727@stud.uni-stuttgart.de}
}

\maketitle

\begin{abstract}
Federated learning (FL) under differential privacy (DP) is typically evaluated on large client cohorts, where subsampling and cross-client averaging amplify privacy and smooth injected noise. Personal biosignal applications such as wearable EEG instead involve small federations of a handful of subjects, a regime whose interaction with formal client-level DP is largely uncharacterized. We study differentially private federated class-incremental learning in this small-cohort regime, combining a public EEG dataset with a custom four-channel MUSE~2 recording set into a $K{=}5$-client federation, and give a masked per-class training protocol that avoids degenerate gradients and missing negative supervision under class-incremental streaming. We propose \textbf{FedEEG}, combining Synaptic-Intelligence consolidation with a lagged adaptive FedProx term, and show it reduces catastrophic forgetting relative to FedAvg, FedProx, and a federated Learning-without-Forgetting baseline; on-device balanced-batch replay reduces mean forgetting by a further 56.7\%. A class-scoping diagnosis yields $50.5\%\pm3.6\%$ accuracy with 6 seeds and $2.02\times$ chance. We also introduce \emph{head-scoped client-level DP}, which protects the classification head and alone reduces the achievable $\varepsilon$ by $200\times$ at a matched round budget; combined with fewer communication rounds it reaches $\varepsilon\!\approx\!10.4$ with utility not statistically distinguishable from non-private training. 
\end{abstract}

\begin{IEEEkeywords}
Federated learning, continual learning, class-incremental learning, differential privacy, EEG, mental state classification.
\end{IEEEkeywords}

\section{Introduction}
Federated learning (FL) under differential privacy (DP) is commonly formulated as optimization over a fixed, known label space \cite{mcmahan2017communication,li2020federated,karimireddy2020scaffold}. Many realistic systems, however, encounter \emph{class-incremental} conditions: new concepts or behaviors appear over time, and models must remain plastic enough to incorporate them while preserving knowledge of earlier classes \cite{parisi2019continual,van2019three}. Combined with client-level DP \cite{geyer2017differentially,mcmahan2018learning,abadi2016deep} and non-IID data \cite{zhao2018federated,li2020federated}, this compounds catastrophic forgetting through gradient noise and cross-client distributional drift.

Existing federated methods, including FedAvg \cite{mcmahan2017communication} and FedProx \cite{li2020federated}, assume fixed-task objectives and do not address interference from sequentially emerging classes. Continual-learning techniques that mitigate forgetting, such as Elastic Weight Consolidation (EWC) \cite{kirkpatrick2017overcoming} and Synaptic Intelligence (SI) \cite{zenke2017continual}, in turn rely on centralized data access or noise-free importance estimates, assumptions violated under noisy, privacy-preserving federated updates.

We study \emph{differentially private federated class-incremental learning} as a regime in which three sources of instability: sequential label-space expansion, cross-client distribution shift, and privacy-induced gradient noise interact and jointly amplify forgetting. Crucially, we target this regime at the client-cohort sizes realistic for personal biosignal applications. Client-level DP is almost always characterized on cohorts of hundreds to millions of participants, where Poisson subsampling gives substantial privacy amplification and cross-client averaging smooths injected noise; a wearable-EEG federation, by contrast, realistically spans a handful of devices, e.g., the members of a single research cohort, care facility, or household, rather than a web-scale user base, and $K{=}5$ is chosen to reflect this deployment scale directly rather than as a matter of convenience. This small-cohort, resource-constrained regime recurs across FL for real-time multimodal sensing \cite{gul2025fedmultiemo}, explicitly-synchronized federated rounds \cite{syncfed}, and in-vehicle personalization \cite{comfortreview}, underscoring the need for methods that remain sound at the cohort sizes real devices actually operate at, not just at web-scale. To make this concrete, each client's raw device stream (sampling rate, electrode montage, ADC resolution) is normalized into a common windowed representation before entering the learning pipeline, decoupling the federated objective from device-specific acquisition parameters. We instantiate this regime with a $K{=}5$ federation combining a public Muse-based EEG dataset \cite{bird2018study} with a custom four-channel MUSE~2 recording set, and make these contributions:

\begin{itemize}
\item \textbf{A degeneracy-avoiding training protocol:} We identify two failure modes of masked-softmax classification under class-incremental streaming: near-zero gradients under single-class increments, and no negative supervision for previously-learned classes when training only on new-class data, and give a masked binary-cross-entropy (BCE) scheme with grouped class increments that avoids both.
\item \textbf{An internally consistent consolidation objective:} We give an SI-based path-integral formulation with the sign convention required for non-negative importance under gradient descent, and a proximal coefficient lagged by one communication round to avoid a circular dependency between a round's update and its own regularization strength.
\item \textbf{Validation against a knowledge-distillation baseline:} On the $K{=}5$ multi-subject, multi-source federation, SI-based consolidation reduces the Forgetting Measure (FM) relative to FedAvg, FedProx, and a calibrated Federated Learning-without-Forgetting (LwF) baseline \cite{li2017learning}, consistently across independent seeds; a further, mechanistically-diagnosed on-device balanced-replay mechanism reduces FM by a further 56.7\%. 
\item \textbf{A class-scoping diagnosis and the primary classification result:} We diagnose why one class underperforms (zero cross-subject data compounded by within-group class competition) and show that scoping it out, rather than accepting the degraded result or discarding classes indiscriminately, yields $50.5\%\pm3.6\%$ accuracy over 6 seeds ($2.02\times$ chance). 
\item \textbf{Head-scoped differential privacy and its limits:} Restricting the formal client-level DP guarantee to the classification head reduces the achievable $\varepsilon$ by $200\times$ on its own, at a matched round budget, isolating this effect from the well-known fact that DP composes across rounds. A further, independent reduction in communication rounds brings $\varepsilon$ down to $\mathbf{10.4}$ overall, with utility not statistically distinguishable from non-private training at our sample size. 
\end{itemize}

\section{Related Work}
We situate FedEEG relative to three overlapping bodies of work: federated optimization under non-IID data, continual and federated continual learning, and differential privacy in federated and continual settings, before stating explicitly where their combined assumptions break down and how FedEEG is positioned relative to the closest alternatives.
\subsection{Federated Learning under Non-IID Distributions}
FL enables distributed training without centralizing data \cite{mcmahan2017communication}. FedAvg \cite{mcmahan2017communication} averages locally-trained models weighted by local sample count; under non-IID client data this can drift away from any single client's optimum, motivating FedProx \cite{li2020federated}, which anchors local updates to the global model via a proximal term (the mechanism we build on in Sec.~\ref{sec:method}), and SCAFFOLD \cite{karimireddy2020scaffold}, which corrects client drift with per-client control variates. All three assume a fixed, stationary label space and provide no mechanism for catastrophic forgetting when labels are introduced sequentially.

\subsection{Continual and Federated Continual Learning}
Continual learning (CL) studies how models acquire new tasks while retaining past knowledge \cite{parisi2019continual,van2019three}. EWC \cite{kirkpatrick2017overcoming} estimates parameter importance via diagonal Fisher information computed once per task boundary; SI \cite{zenke2017continual}, which we build on, instead accumulates importance \emph{online} via a path integral of per-step loss reduction, avoiding a separate estimation pass. Knowledge-distillation methods such as LwF \cite{li2017learning} constrain a model's outputs on new data to match a frozen copy of the previous model; we use LwF as an empirical baseline (Sec.~\ref{sec:results}). All of these are developed for centralized, noise-free settings. Extending CL to FL introduces client-drift and communication constraints: FedWeIT \cite{yoon2021federated} decomposes parameters into shared and sparse client-specific components, Dong et al.\ \cite{dong2021federated} use class-aware gradient compensation with cross-client distillation, and TARGET \cite{zhang2023federated} trains a generator to synthesize old-class data instead of storing exemplars. These assume relaxed or absent privacy constraints and centralized-style importance or generator training.

\subsection{Differential Privacy in Federated and Continual Learning}
DP provides formal guarantees against information leakage by perturbing gradients or updates \cite{abadi2016deep,geyer2017differentially,mcmahan2018learning}, almost always in fixed-task FL. Only recently has DP been studied \emph{jointly} with continual learning: Ouyang et al.~\cite{dpfcl2023} present a systematic study of DP in federated continual learning and find several gradient-based CL strategies become impractical once combined with DP noise, consistent with what we find independently before showing a partial remedy (Sec.~\ref{sec:results}). A separate line on \emph{scoped}/\emph{sparsified} DP improves the FL-DP privacy--utility trade-off by reducing the effective dimensionality of the privatized update: Hu et al.\ \cite{hu2023sparsified} sparsify the perturbed update directly, while non-federated DP fine-tuning shows that privatizing only a network's last layer can suffice for good utility \cite{tramer2021differentially}. To our knowledge, no prior work combines importance-weighted consolidation with a layer-scoped DP mechanism for federated class-incremental learning, nor characterizes this combination in the small-cohort ($K\!\le\!10$) regime typical of personal biosignal federations.

Existing FL, CL, and DP-in-FL work rests on mutually incompatible assumptions (fixed label space; centralized, noise-free gradients; fixed-task regimes), so the coupled regime we study falls outside all three. Relative to rehearsal-based CL, storing and replaying past samples under client-level DP either requires sensitive on-device buffers or noise that degrades gradient quality; FedEEG instead encodes past knowledge in parameter-wise importance, and our balanced replay (Sec.~\ref{sec:replay}) keeps retained data on-device, outside the DP guarantee's scope. Relative to EWC-style offline penalties, SI's online path integral adapts to irregular increment boundaries, participation, and per-client class support.

\section{FedEEG: Privacy-Preserving Federated Class-Incremental Learning}
\label{sec:method}
This section defines the class-incremental problem (\ref{sec:problem}), the combined stability--plasticity objective (\ref{sec:objective}), the client-level DP mechanism and our head-scoped variant (\ref{sec:dp}), and on-device personalization and its privacy scope (\ref{sec:personalization}).

\subsection{Problem Definition}
\label{sec:problem}
We consider an FL system with $K$ clients. Client $k$ observes an online stream $\{(x_t^k,y_t^k)\}_{t=1}^{\infty}$ from a non-stationary distribution, whose label space expands according to $\mathcal{Y}_1\subset\mathcal{Y}_2\subset\cdots\subset\mathcal{Y}_T$. At increment $t$, samples belong to $\mathcal{Y}_t$, but the learner must retain performance on all previously observed classes without access to past raw data, under $(\varepsilon,\delta)$-DP with respect to a single client's participation and data. 

\subsection{Unified Continual Federated Objective}
\label{sec:objective}
At round $r$, each client $k\in\mathcal{K}_r$ initializes $\theta_k\leftarrow\theta^{r-1}$ and minimizes
\begin{equation}
\mathcal{L}_{\text{FedEEG}}(\theta_k)
= \ell_{\mathrm{CE}}(f_{\theta_k}(x), y)
+ \alpha_k^{(r)} \mathcal{R}_{\mathrm{SI}}(\theta_k)
+ \beta_k^{(r)} \mathcal{R}_{\mathrm{prox}}(\theta_k),
\label{eq:fedEEG_loss_SI}
\end{equation}
where $\mathcal{R}_{\mathrm{SI}}(\theta_k)=\sum_i\Omega_{k,i}(\theta_{k,i}-\theta_{k,i}^{\mathrm{ref}})^2$ penalizes movement of high-importance parameters away from their last-consolidated value, and $\mathcal{R}_{\mathrm{prox}}(\theta_k)=\|\theta_k-\theta^{r-1}\|_2^2$ penalizes drift from the current global model.

\paragraph{Masked classification loss} We use masked per-class BCE rather than masked softmax:
\begin{equation}
\ell_{\mathrm{CE}}(f_{\theta_k}(x), y) = \frac{1}{|\mathcal{Y}^{\mathrm{seen}}_t|}
\sum_{j \in \mathcal{Y}^{\mathrm{seen}}_t} \mathrm{BCE}\bigl(\sigma(f_{\theta_k}(x)_j),\, \mathbb{1}[y{=}j]\bigr),
\label{eq:masked_bce}
\end{equation}
excluding unseen classes entirely rather than down-weighting them. Masked softmax is degenerate whenever only one class is unmasked (loss $\approx 0$ regardless of model output, no gradient), and even with multiple classes unmasked it gives no negative supervision for previously-seen classes when training on new-class-only data. Masked BCE supplies an implicit ``not class $j$'' target for every other currently-seen class, which is what gives any consolidation mechanism a signal to act on. We verified this directly: an isolated two-class diagnostic under masked softmax showed exactly zero gradient and zero weight movement during single-class training, while masked BCE produced non-zero, SI-strength-dependent movement.

\paragraph{Synaptic importance} Following SI \cite{zenke2017continual}, each client accumulates a path integral over every local step \emph{within the current increment} -- $\omega_{k,i}$ is reset to zero only when a new increment begins, not every communication round, so it integrates over all rounds belonging to that increment:
\begin{equation}
\omega_{k,i} \leftarrow \omega_{k,i} - g_{k,i}^{(t)} \odot (\theta_{k,i}^{(t)} - \theta_{k,i}^{(t-1)}),
\label{eq:si_omega_corrected}
\end{equation}
where $g_{k,i}^{(t)}=\partial\mathcal{L}/\partial\theta_{k,i}$. The leading minus sign is required for $\omega_{k,i}\!\ge\!0$ under gradient descent: since $\theta_{k,i}^{(t)}-\theta_{k,i}^{(t-1)}=-\eta g_{k,i}^{(t)}$, Eq.~\eqref{eq:si_omega_corrected} reduces to $\omega_{k,i}\mathrel{+}=\eta(g_{k,i}^{(t)})^2\ge0$. At the \emph{end} of each increment, this path integral is converted into an importance contribution and \emph{added} to the running total rather than overwriting it:
\begin{equation}
\Omega_{k,i} \leftarrow \Omega_{k,i} + \frac{\omega_{k,i}}{(\theta_{k,i}-\theta^{\mathrm{ref}}_{k,i})^2 + \xi},
\label{eq:si_consolidate_corrected}
\end{equation}
with damping $\xi>0$, after which $\theta_{k,i}^{\mathrm{ref}}\leftarrow\theta_{k,i}$ marks the new reference point for the \emph{next} increment's displacement penalty. The additive update in Eq.~\eqref{eq:si_consolidate_corrected} is what makes this Synaptic Intelligence rather than a one-increment trust region: importance consolidated while learning $\{$neutral, relaxed$\}$ is preserved in $\Omega_{k,i}$ when the client later moves on to $\{$concentrated, alerted$\}$ and then $\{$drowsy$\}$, rather than being discarded each time the reference point advances.

\paragraph{Adaptive federated alignment} $\beta_k^{(r)}$ anchors clients with small recent updates more strongly and clients with large recent updates (e.g., adapting to a new class) more loosely. It is lagged by one round to remove a circular dependency on the not-yet-available current-round update:
\begin{equation}
\beta_k^{(r)} = \beta_0 \cdot \exp\!\bigl(-\gamma\,\lVert \tilde{\Delta\theta}_k^{(r-1)} \rVert_2\bigr),
\quad \tilde{\Delta\theta}_k^{(0)} := 0,
\label{eq:beta_lag_corrected}
\end{equation}
using client $k$'s already-transmitted update norm from round $r{-}1$ -- this is genuinely round-level, since drift is a round-level phenomenon. The SI strength, by contrast, grows with how much importance has accumulated \emph{across increments} so far, saturating toward its base scale $\alpha_0$ as more is consolidated:
\begin{equation}
\alpha_k^{(r)} = \alpha_0 \cdot \frac{\Omega^{\mathrm{sum}}_k}{\Omega^{\mathrm{sum}}_k+\lambda},
\qquad \Omega^{\mathrm{sum}}_k := \sum_i \Omega_{k,i},
\label{eq:alpha_adapt}
\end{equation}
with $\lambda>0$ preventing unbounded growth. Because $\Omega_k$ is only updated at increment boundaries (Eq.~\eqref{eq:si_consolidate_corrected}), $\alpha_k^{(r)}$ is constant across all rounds within the same increment and steps up only when a new increment begins; $\beta_k^{(r)}$ updates every round. Both coefficients are computed from already-available information only -- $\alpha_k^{(r)}$ from the previous increment's consolidated $\Omega_k$, $\beta_k^{(r)}$ from the previous round's transmitted update -- so neither depends on the very update it is regularizing.

\subsection{Privacy-Preserving Federated Optimization}
\label{sec:dp}
Each client transmits a clipped, noised version of its raw update $\Delta\theta_k^r=\theta_k^r-\theta^{r-1}$:
\begin{equation}
\tilde{\Delta\theta}_k^r = \mathrm{clip}(\Delta\theta_k^r, S) + \mathcal{N}(0,\sigma^2 S^2 I_d),
\label{eq:dp_mechanism}
\end{equation}
with $\mathrm{clip}(v,S)=v/\max(1,\|v\|_2/S)$. This is \emph{update-level} (output) perturbation: noise is added once per round to the finished local update, not to individual SGD steps, so it must be composed via a per-round R\'enyi-DP accountant rather than a per-step DP-SGD bound. The stated guarantee presumes secure aggregation, so the server observes only the noised sum of client updates each round; without it, the mechanism yields only a weaker, non-aggregated guarantee. Unless otherwise noted, all clients participate in every round ($q{=}1$, no subsampling) for the main results; Sec.~\ref{sec:dp_results} separately reports what happens under partial participation. All $\varepsilon$ values we report are the total, composed guarantee over the full $R$-round training run, converted from an $(\alpha,\rho)$-RDP composition to $(\varepsilon,\delta)$-DP at the standard $\delta=10^{-5}$.

\textit{Head-scoped DP:} Let $\theta=(\theta^{\mathrm{feat}},\theta^{\mathrm{head}})$ partition the model into a feature extractor ($d^{\mathrm{feat}}$ parameters) and classification head ($d^{\mathrm{head}}\ll d^{\mathrm{feat}}$; 325 vs.\ 23,584 in our architecture). We restrict the mechanism to the head:
\begin{equation}
\tilde{\Delta\theta}_k^{\mathrm{head}} = \mathrm{clip}(\Delta\theta_k^{\mathrm{head}}, S) + \mathcal{N}(0,\sigma^2 S^2 I_{d^{\mathrm{head}}}),
\label{eq:head_scoped_dp}
\end{equation}
\begin{equation}
\tilde{\Delta\theta}_k^{\mathrm{feat}} = \Delta\theta_k^{\mathrm{feat}} \quad \text{(sent unmodified).}
\label{eq:head_scoped_dp_feat}
\end{equation}
$(\varepsilon,\delta)$ depends on $\sigma$ via RDP composition but the noise magnitude needed to hold a fixed noise-to-signal ratio $r=\sqrt{d}\,\sigma$ scales with $\sqrt{d}$; restricting $d$ to $d^{\mathrm{head}}$ therefore permits a $\sqrt{(d^{\mathrm{feat}}+d^{\mathrm{head}})/d^{\mathrm{head}}}$-fold larger $\sigma$ at the same $r$, hence smaller $\varepsilon$ at matched utility-relevant noise. \emph{The formal guarantee covers only $\theta^{\mathrm{head}}$}; the feature extractor is federated but not formally DP under this scheme.

This scope limitation is not merely a technicality: $\theta^{\mathrm{feat}}$ is trained on private EEG data and, absent DP noise, its update remains exposed to standard FL attacks (gradient inversion, membership inference) that the formal guarantee does not bound. Two factors partially mitigate this: the update is aggregated under the secure-aggregation assumption above, so no individual $\Delta\theta_k^{\mathrm{feat}}$ is exposed to the server; and the feature extractor is small (23{,}584 parameters), which is believed to reduce memorization capacity. We recommend that deployments either (a) pair head-scoped DP with a stronger non-DP defense on the feature path (e.g., cryptographic or trusted-execution secure aggregation beyond our honest-but-curious assumption), or (b) extend DP to the full model, accepting the utility cost quantified in Sec.~\ref{sec:dp_results}. Head-scoped DP is thus a specific, stated point on this trade-off, not a substitute for full-model protection where the threat model requires it.

\subsection{Subject-Level Personalization}
\label{sec:personalization}
The global classification head can optionally be fine-tuned on-device using a local calibration buffer $\mathcal{B}_k$. The resulting personalized parameters $\theta_k^{\mathrm{pers}}$ are never transmitted and therefore fall outside the client-level DP guarantee established above, which covers only quantities that leave the client. We specify this mechanism and its privacy scope here; a systematic empirical evaluation of subject-level personalization under the $K{=}5$ protocol is left to future work. Algorithm~\ref{alg:fedEEG} summarizes the full procedure.

\begin{algorithm}[t]
\footnotesize
\caption{FedEEG training and privatized communication}
\label{alg:fedEEG}
\begin{algorithmic}[1]
\Require Clients $K$; increments $n{=}1,\dots,N$ with rounds-per-increment $R_n$ and newly-introduced classes $\mathcal{Y}_n$; step size $\eta$; SI damping $\xi$; clip norm $S$; noise multiplier $\sigma$; base scales $\alpha_0,\beta_0$; decay $\gamma$; normalizer $\lambda$
\Ensure Final global model $\theta$
\State Init.\ $\theta$; $\forall k$: $\Omega_k\!\leftarrow\!0$, $\theta_k^{\mathrm{ref}}\!\leftarrow\!\theta$, $\tilde{\Delta\theta}_k^{(\mathrm{prev})}\!\leftarrow\!0$; $\mathcal{Y}^{\mathrm{seen}}\leftarrow\emptyset$
\For{$n = 1,\dots,N$} \Comment{increment loop}
  \State $\mathcal{Y}^{\mathrm{seen}} \leftarrow \mathcal{Y}^{\mathrm{seen}}\cup\mathcal{Y}_n$
  \State $\forall k$: $\omega_k \leftarrow 0$ \Comment{reset only here, not every round}
  \For{$r = 1,\dots,R_n$} \Comment{round loop within increment $n$}
    \State Server selects $\mathcal{K}_r$, broadcasts $\theta$
    \For{each $k\in\mathcal{K}_r$ \textbf{in parallel}}
      \State $\theta_k \leftarrow \theta$
      \State Compute $\alpha_k^{(r)}$ \eqref{eq:alpha_adapt} from $\Omega_k$; $\beta_k^{(r)}$ \eqref{eq:beta_lag_corrected} from $\tilde{\Delta\theta}_k^{(\mathrm{prev})}$
      \For{each local step $t$ on masked-BCE batches over $\mathcal{Y}^{\mathrm{seen}}$ \eqref{eq:masked_bce}}
        \State $g_k^{(t)} \leftarrow \nabla_{\theta_k}\mathcal{L}_{\text{FedEEG}}$ \eqref{eq:fedEEG_loss_SI} \Comment{uses $\Omega_k$ from increment $n{-}1$}
        \State $\theta_k \leftarrow \theta_k - \eta\, g_k^{(t)}$
        \State $\omega_k \leftarrow \omega_k - g_k^{(t)}\odot(\theta_k^{(t)}-\theta_k^{(t-1)})$ \Comment{\eqref{eq:si_omega_corrected}, accumulates over all of $n$'s rounds}
      \EndFor
      \State $\Delta\theta_k \leftarrow \theta_k - \theta$
      \State $\tilde{\Delta\theta}_k \leftarrow$ Eq.~\eqref{eq:dp_mechanism} (or head-scoped \eqref{eq:head_scoped_dp}--\eqref{eq:head_scoped_dp_feat})
      \State Transmit $\tilde{\Delta\theta}_k$; $\tilde{\Delta\theta}_k^{(\mathrm{prev})} \leftarrow \tilde{\Delta\theta}_k$
    \EndFor
    \State $\theta \leftarrow \theta + \sum_{k\in\mathcal{K}_r} \frac{\sqrt{n_k}}{\sum_{j}\sqrt{n_j}}\, \tilde{\Delta\theta}_k$
  \EndFor
  \State $\forall k$: $\Omega_k \leftarrow \Omega_k + \omega_k \oslash \bigl((\theta_k-\theta_k^{\mathrm{ref}})^2+\xi\bigr)$ \Comment{\eqref{eq:si_consolidate_corrected}, end of increment $n$ only}
  \State $\forall k$: $\theta_k^{\mathrm{ref}} \leftarrow \theta_k$
\EndFor
\State \textbf{return} $\theta$
\end{algorithmic}
\end{algorithm}

\section{Experimental Setup}
\label{sec:setup}
This section specifies everything needed to reproduce the results in Sec.~\ref{sec:results}: the federated data partition and class-introduction schedule, the model architecture, the evaluation metrics (and why three are reported together), and the hyperparameters together with how they were calibrated.
\subsection{Datasets and Federated Partitioning}
We evaluate FedEEG with $K{=}5$ clients from two sources. Four clients ($k{=}0..3$) are the four distinct subjects of the public Bird et al.\ dataset \cite{bird2018study}, each contributing raw four-channel (TP9, AF7, AF8, TP10) Muse EEG for the neutral/relaxed/concentrating classes only. The fifth client is an in-house pilot dataset recorded with a consumer-grade MUSE~2 headband under the same four-channel montage, pooled across its 29 recording sessions and contributing all five classes; since the federation treats this source as a single client throughout, per-session subject identity does not affect any result in this paper other than the dual-benchmark check in Sec.~\ref{sec:primary}, where we already report this pooled source separately from the real-subject Bird-only federation. Four of five clients are therefore entirely missing two of the five classes, a realistic, non-arbitrary property of the merged data rather than a partition artifact. Train/test splits are performed at the \emph{file} (recording-session) level, not the window level, since windows overlap 50\% (stride 128, window 256) and a window-level split would let near-duplicate, temporally adjacent windows leak across the split. Classes are introduced in grouped increments, \{neutral, relaxed\} $\to$ \{concentrated, alerted\} $\to$ \{drowsy\}, rather than one at a time (Sec.~\ref{sec:objective}); this order and grouping was fixed prior to running any class-incremental experiment and was not selected based on observed results. 

\subsection{Model Architecture}
We use a compact 1D CNN appropriate for on-device federated training: three convolutional layers (kernel sizes $\{7,5,3\}$, widths $\{32,64,64\}$), global average pooling, and a linear head, totaling $23{,}909$ scalar parameters ($d^{\mathrm{feat}}{=}23{,}584$, $d^{\mathrm{head}}{=}325$), which directly determines the head-scoping dimensionality reduction used in Secs.~\ref{sec:dp} and \ref{sec:dp_results}.

\subsection{Evaluation Metrics}
Let $R_{t,c}$ be test accuracy on class $c$ after increment $t$, $T$ the number of increments, and $\mathrm{intro}(c)$ the increment at which $c$ is introduced. We report:
\begin{equation}
\mathrm{ACC} = \tfrac{1}{C}\textstyle\sum_{c=1}^{C} R_{T-1,c},
\end{equation}
\begin{equation}
\mathrm{BWT} = \tfrac{1}{C}\textstyle\sum_{c=1}^{C}\bigl(R_{T-1,c} - R_{\mathrm{intro}(c),c}\bigr),
\end{equation}
\begin{equation}
\mathrm{FM} = \tfrac{1}{C}\textstyle\sum_{c=1}^{C} \Bigl(\max_{t\in[\mathrm{intro}(c),T-1]} R_{t,c} - R_{T-1,c}\Bigr).
\end{equation}
ACC alone is uninformative for a degenerate model: a classifier that always predicts one class scores $1/C$ on it. Under the main 9-round class-incremental protocol (Table~\ref{tab:main_results_revised}), a class's accuracy is highest immediately after introduction and does not recover, so $\mathrm{FM}=-\mathrm{BWT}$ there; we report FM (non-negative, ``larger = more forgetting'') for that table. This is a property of the observed accuracy trajectories, not of the definitions in general: FM $\ge 0$ always by construction, whereas BWT can be positive if a class's accuracy is higher at the end of training than immediately after it was introduced. That is precisely what happens in the reduced-round ablation of Table~\ref{tab:dp_results}: with only three total rounds and local epochs raised sixfold per round to compensate, an early-introduced class need not yet be at its own local optimum at the moment of introduction, so continued training on later increments can still lift it before the run ends, giving a positive mean BWT. We therefore report BWT and FM independently rather than via the identity for that ablation, which is itself a useful observation: the round/epoch schedule, not just the consolidation method, shapes whether forgetting curves are monotonic.

\subsection{Hyperparameters}
Table~\ref{tab:hyperparams} lists all hyperparameters used to produce the main results. $\alpha_0,\beta_0$ (and, for the LwF baseline, $\lambda_{\mathrm{kd}}{=}10$) were calibrated via an isolated two-class diagnostic (train class $A$ only, then $B$ only, measure retention of $A$) on training-split data disjoint from the test folds used for every reported ACC/BWT/FM number, run \emph{before} the main comparison, which showed a sharp threshold near $\alpha_0\!\approx\!20$ between ``forgets $A$ completely'' and ``cannot learn $B$ at all''.

\begin{table}[t]
\scriptsize
\setlength{\tabcolsep}{3.5pt}
\caption{Hyperparameters used in all reported experiments.}
\label{tab:hyperparams}
\centering
\begin{tabular}{ll}
\toprule
Parameter & Value \\
\midrule
Window / stride & 256 / 128 samples (50\% overlap) \\
Rounds/increment, local epochs (main) & 3, 2 \\
Batch size, optimizer & 64, Adam ($\eta{=}10^{-3}$, reset/round) \\
Aggregation weighting & $\sqrt{n}$-weighted (Sec.~\ref{sec:primary}) \\
SI: $\alpha_0,\lambda,\xi$ & 25.0, 1.0, $10^{-3}$ \\
Prox: $\beta_0,\gamma$ & 5.0, 1.0 \\
DP clip norm $S$ & 5.0 \\
Split & file-level, 80/20 within source \\
Federation & $K{=}5$: 4 Bird subj.\ + 1 pooled custom \\
Seeds & 0--3 (0--5 for headline result, Table~\ref{tab:classscoping}) \\
\bottomrule
\end{tabular}
\end{table}

\section{Results}
\label{sec:results}
We first ask how well this federation classifies EEG mental states under standard (non class-incremental) training (\ref{sec:primary}), since the class-incremental experiments that follow (\ref{sec:forgetting}) are a stress test of forgetting dynamics and should not be mistaken for the model's classification ceiling. We then report the privacy--utility trade-off (\ref{sec:dp_results}).

\subsection{Primary Classification Performance}
\label{sec:primary}
We report classification performance in three steps: the choice of aggregation weighting and why it matters at this federation's extreme client-size imbalance, the resulting per-class accuracy and the class-scoping diagnosis it motivates, and validation against two independent single-source benchmarks.

\begin{figure}[t]
\centering
\includegraphics[width=\linewidth]{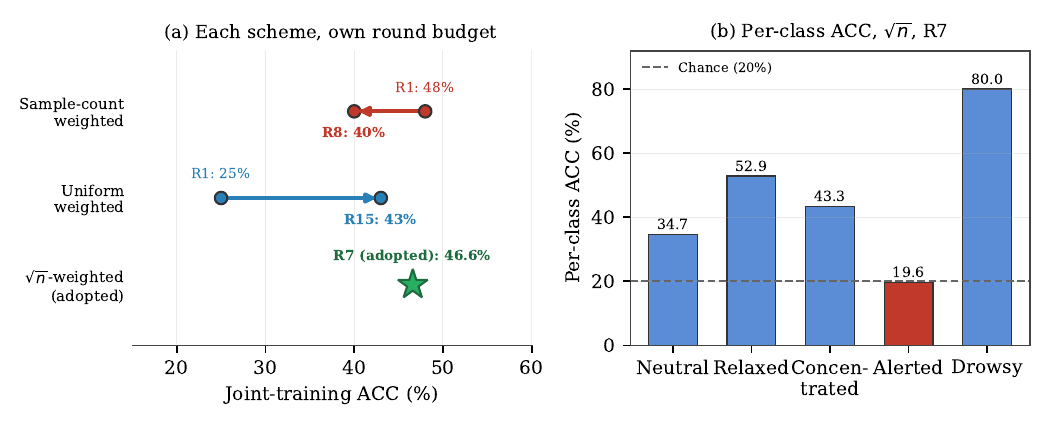}
\caption{(a) Each aggregation scheme shown at its own evaluated rounds, not on a shared round axis: sample-count-weighted FedAvg \emph{degrades} from round 1 to round 8 under this federation's extreme client-size imbalance, uniform weighting \emph{improves} from round 1 to round 15 but dilutes the two client-unique classes (Sec.~\ref{sec:primary}), and $\sqrt{n}$-weighting is the adopted operating point at its own near-peak (round 7). (b) Per-class accuracy under $\sqrt{n}$-weighted joint training at round 7: every class substantially exceeds chance except ``alerted,'' motivating the class-scoping diagnosis in Table~\ref{tab:classscoping}.}
\label{fig:joint_training}
\end{figure}

\paragraph{Aggregation weighting matters} With sample-count-weighted FedAvg, accuracy \emph{degrades} with more rounds (round~1: 47--49\%, round~8: 40\%), since the pooled custom-dataset client holds $\sim$85\% of samples and increasingly dominates the average. Uniform per-client weighting stabilizes this (25\%$\to$43\% over 15 rounds) but drives the two classes unique to that client toward zero accuracy. We adopt $\sqrt{n}$-weighting as a middle ground (Fig.~\ref{fig:joint_training}a).

At $\sqrt{n}$-weighting, 7 rounds, joint (non-incremental) training over all 5 classes reaches $\mathbf{46.6\%\pm4.5\%}$ (4 seeds) against 20\% chance ($2.33\times$); per-class accuracy (Fig.~\ref{fig:joint_training}b) is well above chance for neutral (34.7\%), relaxed (52.9\%), concentrated (43.3\%), and drowsy (80.0\%), but \emph{alerted} sits at chance ($19.6\%$, one seed at $0\%$). We diagnosed this after observing it, not before: the public dataset contributes zero alerted/drowsy samples at all, so both classes depend entirely on the single custom-dataset client; unlike drowsy, alerted additionally shares its increment group with concentrated, which outnumbers it $\sim\!4.7{:}1$ even within that one client's data, compounding cross-subject starvation with within-group competition.

\begin{table}[t]
\scriptsize
\setlength{\tabcolsep}{3pt}
\caption{Joint federated training under three class-scoping configurations, reported in the order they were run. 4-class (drop only ``alerted'') attains the highest accuracy while keeping every retained class well above its own chance level.}
\label{tab:classscoping}
\centering
\begin{tabular}{lccc}
\toprule
Configuration & Classes & Seeds & ACC (chance$\times$) \\
\midrule
5-class (full) & all 5 & 4 & $46.6\%\pm4.5\%$ (2.33$\times$) \\
3-class (shared) & neu., rel., conc. & 4 & $46.5\%\pm3.6\%$ (1.40$\times$) \\
4-class (drop alerted) & neu., rel., conc., drow. & 6 & $\mathbf{50.5\%\pm3.6\%}$ (\textbf{2.02$\times$}) \\
\bottomrule
\end{tabular}
\end{table}

This diagnosis, made after inspecting the 5-class result, yields a testable prediction: dropping only alerted (not drowsy) should recover accuracy without discarding an already-working class. Table~\ref{tab:classscoping} confirms it: reporting all three configurations we ran, in order, rather than only the best one. The 4-class configuration reaches $50.5\%\pm3.6\%$ (95\% CI $[46.4,54.7]\%$, $t$-distribution). This configuration uses 6 seeds rather than the 4 used elsewhere: seeds 0--3 were run under the same protocol as every other configuration and already showed the same qualitative result, and seeds 4--5 were added afterward, with none excluded at any point, specifically to tighten the confidence interval on our headline number. Every retained class is substantially above chance (neutral $36.1\%\pm8.6\%$, relaxed $58.2\%\pm12.6\%$, concentrated $46.6\%\pm13.3\%$, drowsy $77.6\%\pm20.9\%$). We treat this as our most defensible classification result, but emphasize its \emph{exploratory, post-hoc} status: the class-scoping decision followed inspection of the 5-class result rather than being pre-registered, so it should be read as a diagnosed and confirmed hypothesis rather than an independently validated outcome; validating the 4-class configuration under a pre-specified protocol or an independent evaluation cohort is the natural next step. In practical terms, $50.5\%$ over four classes is sufficient to distinguish coarse operating states (e.g., flagging a likely drowsy vs.\ engaged state for a downstream alerting threshold) but falls well short of what would be required for fine-grained, clinical-grade mental-state discrimination. The class-incremental experiments below (Sec.~\ref{sec:forgetting}) deliberately retain the full 5-class space, since alerted performing poorly is itself informative there (a class can be both poorly learned \emph{and} subject to catastrophic forgetting); re-running Sec.~\ref{sec:forgetting} under 4-class scoping is the most direct extension of this work.

\paragraph{Validation on two independent open benchmarks} As a complementary check, we evaluate each open source alone. \emph{Bird-only} ($K{=}4$ real, distinct subjects, 3 classes) reaches $62.3\%\pm2.4\%$ (chance 33.3\%, $1.87\times$), consistently across all 4 seeds. \emph{Custom-only} (5 classes, $K{=}3$ Dirichlet pseudo-clients, since the pilot source is pooled rather than collected per-subject) reaches a higher mean, $46.3\%\pm13.7\%$ (chance 20\%, $2.32\times$), but with an order of magnitude higher variance, including one seed at $30.1\%$ with three of five classes collapsing to $0\%$. The genuinely multi-subject federation is not only more accurate but markedly more reliable across seeds than the pseudo-client one, direct evidence that per-subject data collection, not just per-file pooling, matters empirically for federated reliability, and that the extreme, single-dominant-client imbalance of our main $K{=}5$ federation (Sec.~\ref{sec:setup}A) is a deliberately severe stress test rather than a representative deployment target; more evenly distributed cohorts should be expected to behave closer to the Bird-only benchmark's low-variance regime than to the pooled-client one.

\subsection{Forgetting Reduction under Class-Incremental Streaming}
\label{sec:forgetting}
This subsection switches to the harder class-incremental protocol (Sec.~\ref{sec:objective}) and compares SI-based consolidation against three baselines, then diagnoses and closes the remaining gap with on-device balanced replay. FedAvg and FedProx are not continual-learning methods, so we add \textbf{Federated LwF} \cite{li2017learning}: at each new increment, the frozen previous-round global model's soft predictions on the same new-class batch are distilled into the current model for all previously-seen classes, using the same BCE convention as Eq.~\eqref{eq:masked_bce}.

\begin{table}[t]
\scriptsize
\setlength{\tabcolsep}{3.5pt}
\caption{ACC/BWT/FM under federated class-incremental EEG streams, $K{=}5$, mean$\pm$std over 4 seeds. $\mathrm{FM}=-\mathrm{BWT}$ under this (9-round) protocol's monotonic decline (Sec.~\ref{sec:results}); see Table~\ref{tab:dp_results} for a shorter-schedule ablation where this does not apply.}
\label{tab:main_results_revised}
\centering
\begin{tabular}{lccc}
\toprule
Method & ACC & BWT & FM \\
\midrule
FedAvg (CE) & $0.200\pm.000$ & $-0.434\pm.032$ & $0.434\pm.032$ \\
CE + Prox & $0.200\pm.001$ & $-0.399\pm.006$ & $0.399\pm.006$ \\
Federated LwF \cite{li2017learning} & $0.200\pm.000$ & $-0.393\pm.012$ & $0.393\pm.012$ \\
CE + Prox + SI & $0.203\pm.004$ & $-0.333\pm.077$ & $0.333\pm.077$ \\
+ Balanced Replay (\ref{sec:replay}) & $0.194\pm.018$ & $\mathbf{-0.144\pm.118}$ & $\mathbf{0.144\pm.118}$ \\
\bottomrule
\end{tabular}
\end{table}

ACC sits at the 5-class chance floor (0.200) for essentially every method, since a model that always predicts the most recently introduced class scores exactly $1/5$; we emphasize this explicitly because it means the contribution of this subsection is mitigation of catastrophic forgetting under a difficult small-cohort setting, not a claim that whole-task class-incremental classification is solved, even though SI measurably slows \emph{how fast} accuracy is lost after each class is introduced. SI-based consolidation reduces mean FM by 23.3\% relative to FedAvg-only and 16.6\% relative to Prox-alone, consistently in the same direction across all four seeds (one-sided sign-test $p{=}0.0625$ against each baseline); paired $t$-tests give $t{=}2.02,p{=}0.137$ and $t{=}1.58,p{=}0.212$ respectively -- not significant at this sample size, and we do not claim significance. Because $n{=}4$ seeds gives a paired test very little power, we additionally report the standardized mean difference (Hedges' $d_{\mathrm{av}}$, computed from the reported group means/SDs rather than per-seed differences, so it is a conservative approximation of the true paired effect when outcomes are positively correlated across seeds, as expected here since all methods share the same seeds and federation): $d_{\mathrm{av}}{=}1.71$ vs.\ FedAvg and $d_{\mathrm{av}}{=}1.21$ vs.\ Prox, both conventionally ``large'' (\,$|d|{>}0.8$\,). We read this as a large, directionally consistent effect that our sample size is underpowered to certify at $p{<}0.05$, not as a null result. Against Federated LwF, SI reduces mean FM by 15.3\% ($t{=}1.51,p{=}0.228$, $d_{\mathrm{av}}{=}1.09$), winning in 3/4 seeds; the fourth is an effective tie ($\mathrm{FM}{=}0.398$ vs.\ $0.394$). LwF is more \emph{consistent} across seeds (std $0.012$ vs.\ $0.077$) while being consistently worse on average which is a genuine trade-off between mean forgetting and run-to-run reliability and not a one-sided win for SI.

\subsubsection{Balanced-Batch Experience Replay}
\label{sec:replay}
Regularization alone still leaves ACC at the chance floor. A first attempt added a small on-device rehearsal buffer (20--250 windows per seen class, sampled locally, never transmitted, so it does not weaken the DP guarantee on transmitted updates) concatenated with new data before batching; across 4 seeds this did \emph{not} reliably beat SI alone (mean FM $0.337\pm0.059$, $p{=}0.87$, winning only 2/4 seeds). We diagnosed why: for the highest-volume client, naive concatenation-then-shuffle yields an \emph{expected} 2 of 64 replay samples per batch when new data outnumbers the buffer $\sim\!60{:}1$ -- present, but too few to contribute meaningful gradient signal against 62 opposing samples. In order to fix this, replay samples are drawn separately and mixed into every batch at a fixed fraction (0.4) regardless of the raw size ratio. This alone reduces mean FM from 0.333 (SI) to $\mathbf{0.144\pm0.118}$ -- a \textbf{56.7\%} reduction -- winning in 4/4 seeds ($t{=}{-}2.54,p{=}0.085$, $d_{\mathrm{av}}{=}1.90$, a large effect that falls just short of $p{<}0.05$ at $n{=}4$), with per-seed reductions from 8.8\% to 99.3\%. Per-class breakdowns confirm this is genuine multi-class retention (two or more classes simultaneously non-trivial at the final increment) which is not an instance of ``nothing was learned, so nothing was forgotten,'' though we reiterate that ACC itself remains near the chance floor (Table~\ref{tab:main_results_revised}): note that balanced replay reduces \emph{how much} is forgotten and not whether the whole-group task is solved.

We also tested a complementary intervention: oversampling minority classes within a multi-class increment group up to the majority count, on top of balanced replay, and report honestly that it does not generalize: it dramatically improved the single weakest seed ($\mathrm{FM}$: $0.311\!\to\!0.063$) but degraded the other three, for a worse mean overall ($0.219\pm0.115$ vs.\ $0.144$). Balanced-batch replay alone is the configuration we recommend and report as FedEEG's best result.

\subsection{Closing the Privacy--Utility Gap}
\label{sec:dp_results}
Naive full-model DP implies $\varepsilon\!\sim\!10^{5}$ at our model's dimensionality. We test three levers: dimensionality reduction (head-scoping), client subsampling, and reduced communication rounds.

\begin{table}[t]
\scriptsize
\setlength{\tabcolsep}{3.5pt}
\caption{ACC/BWT/FM at the reportable DP operating point ($\varepsilon\approx10.4$, head-scoped, 3 rounds) vs.\ its matched non-private counterpart at the identical 3-round schedule and the original 9-round non-private protocol. Mean$\pm$std, $n{=}4$ seeds.}
\label{tab:dp_results}
\centering
\begin{tabular}{lccc}
\toprule
Configuration & ACC & BWT & FM \\
\midrule
Non-DP, reduced-round (3 rd.) & $0.239\pm.077$ & $0.200\pm.134$ & $0.200\pm.134$ \\
DP, head-scoped, $\varepsilon{\approx}10.4$ & $0.239\pm.076$ & $-0.315\pm.065$ & $0.315\pm.065$ \\
Non-DP, original (9 rd.) & $0.203\pm.004$ & $-0.333\pm.077$ & $0.333\pm.077$ \\
\bottomrule
\end{tabular}
\vspace{2pt}
\par\raggedright\scriptsize\textit{Note:} row 3 reproduces Table~\ref{tab:main_results_revised}'s CE+Prox+SI row (same protocol) for context; the properly matched control for isolating the effect of DP is row 1, at the identical 3-round schedule as row 2. BWT and FM are reported independently rather than via $\mathrm{FM}{=}{-}\mathrm{BWT}$, since row 1's shorter schedule yields positive BWT (Sec.~\ref{sec:results}).
\end{table}

\begin{figure}[t]
\centering
\includegraphics[width=\linewidth]{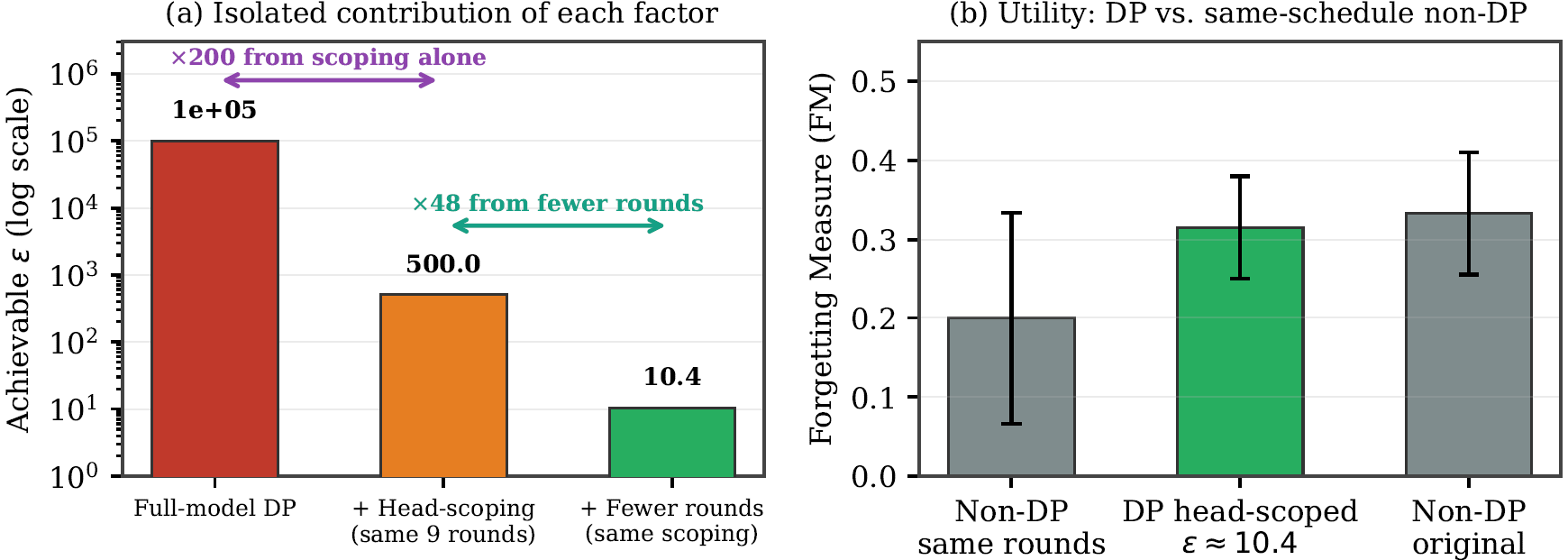}
\caption{(a) The achievable $\varepsilon$ reduction decomposed into two \emph{isolated} factors, each measured while holding the other fixed: head-scoping alone (at the original 9-round budget) already gives $200\times$; a further, independent $48\times$ comes from reducing communication rounds. The larger share of the total $9{,}615\times$ reduction is attributable to head-scoping itself and not to training for fewer rounds. (b) Utility (FM) of the DP-protected 3-round run against its properly matched non-DP counterpart at the identical round budget (paired $t$-test, $p{=}0.20$), and against the original 9-round non-private protocol for context only.}
\label{fig:privacy_utility}
\end{figure}

Neither head-scoping nor client subsampling alone closes the gap to a reportable operating point, holding everything else fixed at the original 9-round schedule. Head-scoping alone reduces $\varepsilon$ from $\sim\!10^{5}$ to $\sim\!500$ at matched noise-to-signal ratio, a $\mathbf{200\times}$ reduction from dimensionality alone, round budget and every other hyperparameter unchanged (Table~\ref{tab:hyperparams}), isolating this mechanism's contribution from the well-known fact that RDP composes additively across rounds, so it cannot be dismissed as ``fewer rounds trivially gives lower $\varepsilon$.'' Subsampling was separately verified with a Poisson-sampled RDP accountant (\texttt{dp\_accounting} \cite{dpaccounting}): at $K{=}5$, dropping participation from $q{=}1.0$ to $q{=}0.2$ reduces $\varepsilon$ by only $\sim$9\%, confirming subsampling amplification is not what is doing the work here. On top of head-scoping, reducing rounds from 9 to 3 (local epochs raised 2$\to$6 to hold total local computation fixed) contributes a further, independent $\mathbf{48\times}$ reduction (500 to 10.4) at $\delta{=}10^{-5}$, cross-checked with a second RDP accountant implementation. The two factors compose to the full $\mathbf{9{,}615\times}$ reduction (Fig.~\ref{fig:privacy_utility}a), with head-scoping responsible for the larger share (58\% on a log scale) -- round-reduction is a real, useful complementary lever and not the primary source of the improvement.

The properly matched comparison for isolating DP's utility cost is between Table~\ref{tab:dp_results}'s rows 1 and 2, both at the identical 3-round schedule: a paired $t$-test per-seed on FM gives $t{=}{-}1.62$, $p{=}0.20$ ($d_{\mathrm{av}}{=}1.06$, a large but, at $n{=}4$, not significant effect), and on ACC gives $t{=}0.05$, $p{=}0.96$ (0.239 vs.\ 0.239, effectively identical). DP incurs a directionally real but not statistically confirmed forgetting cost at matched schedule (row 2 vs.\ row 1), while ACC is unaffected within measurement noise; the comparison against the original 9-round protocol (row 3) is contextual, showing the DP-protected 3-round result is not clearly worse than a substantially longer non-private run, but is not itself the matched control. Table~\ref{tab:dp_results} reports the full picture across all four seeds rather than the single best seed (ACC $=36.3\%$ alone).

Separately, at a \emph{fixed} noise-to-signal ratio, head-scoping does \emph{not} improve utility over full-model DP: concentrating noise entirely on the low-redundancy decision layer degrades classification more than spreading the same ratio across the redundant, SI-consolidated feature extractor, even though the former achieves a far smaller $\varepsilon$ at that ratio. Scoping alone, at matched ratio, is not a free lunch; the gain in this work comes from scoping combined with fewer, higher-compute rounds, which jointly move the achievable operating point into reportable territory in absolute terms. 

The three result threads address distinct questions and should be read together. Sec.~\ref{sec:primary} establishes what this federation can classify under favorable, non-incremental conditions ($50.5\%$ at $2.02\times$ chance); Sec.~\ref{sec:forgetting} shows that under the harder class-incremental protocol, consolidation and balanced replay recover a large share of the retention that regularization alone leaves on the table, without exceeding the whole-group chance floor on ACC; and Sec.~\ref{sec:dp_results} shows a formally private version of the protocol, scoped to the classification head, is achievable at a reportable $\varepsilon$ with a directionally real but not statistically confirmed utility cost at matched schedule. Taken together: the dominant remaining bottleneck is retention capacity under label-space expansion, not privacy noise. Head-scoped DP is close to a small, quantified cost at this cohort size, whereas closing the ACC gap in Table~\ref{tab:main_results_revised} is not.

\section{Conclusion}
We studied differentially private federated class-incremental learning for EEG mental-state classification in the small-cohort regime typical of personal biosignal applications. Using two merged datasets spanning a real four-subject federation and one pooled custom source, we showed that SI-based consolidation with a lagged adaptive proximal coefficient consistently reduces forgetting relative to FedAvg, FedProx, and a knowledge-distillation baseline, and that on-device balanced-batch replay reduces it substantially further without weakening the client-level DP guarantee -- while final accuracy under this streaming protocol remains near the chance floor for every method. An exploratory class-scoping diagnosis yields our best classification result, $50.5\%\pm3.6\%$ over 6 seeds, sufficient for coarse-grained state discrimination but not yet fine-grained classification. Head-scoped DP reduces the achievable privacy budget by $200\times$ on its own; combined with reduced communication rounds it reaches $\varepsilon\!\approx\!10.4$ at a forgetting cost that is directionally present but not statistically confirmed at matched schedule, and this guarantee covers only the classification head, a scope limitation with concrete practical implications we discuss explicitly in Sec.~\ref{sec:dp}. All results are at $K{=}5$; extending to larger cohorts, where subsampling amplification and reduced client dominance both become relevant, and systematically evaluating on-device personalization, are the most direct next steps toward deployable privacy-preserving federated biosignal learning.

\section*{Acknowledgment}
This publication is carried out as part of the Hardware Abstraction Layer for Software Defined Vehicle (HAL4SDV) project, No.\ 101139789, co-funded by the European Union (Chips JU).

\bibliographystyle{IEEEtran}
\bibliography{references}

\end{document}